\documentclass[nohyperref]{article}

\usepackage{microtype}
\usepackage{graphicx}
\usepackage{subfigure}
\usepackage{booktabs} % for professional tables

\usepackage{hyperref}

\usepackage[accepted]{icml2022}

\usepackage{amsmath}
\usepackage{amssymb}
\usepackage{mathtools}
\usepackage{amsthm}

\usepackage[capitalize,noabbrev]{cleveref}

\theoremstyle{plain}

\theoremstyle{definition}

\theoremstyle{remark}

\usepackage[textsize=tiny]{todonotes}

\icmltitlerunning{Robust Augmentation for Multivariate Time Series Classification}

\begin{document}

\twocolumn[
\icmltitle{Robust Augmentation for Multivariate Time Series Classification}



\icmlsetsymbol{equal}{*}

\begin{icmlauthorlist}
\icmlauthor{Hong Yang}{yyy}
\icmlauthor{Travis Desell}{yyy}
\end{icmlauthorlist}

\icmlaffiliation{yyy}{Department of XXX, University of YYY, Location, Country}

\icmlkeywords{Machine Learning, ICML}

\vskip 0.3in
]




\begin{abstract}

Neural networks are capable of learning powerful representations of data, but they are susceptible to overfitting due to the number of parameters. This is particularly challenging in the domain of time series classification, where datasets may contain fewer than 100 training examples. In this paper, we show that the simple methods of cutout, cutmix, mixup, and window warp improve the robustness and overall performance in a statistically significant way for convolutional, recurrent, and self-attention based architectures for time series classification. We evaluate these methods on 26 datasets from the University of East Anglia Multivariate Time Series Classification (UEA MTSC) archive and analyze how these methods perform on different types of time series data.. We show that the InceptionTime network with augmentation improves accuracy by 1\% to 45\% in 18 different datasets compared to without augmentation. We also show that augmentation improves accuracy for recurrent and self attention based architectures. 

\end{abstract}

\section{Introduction}

Time Series Classification (TSC) involves building models to predict a discrete target class from ordered input variables.  Over recent years, new TSC algorithms have made significant improvement over the previous state of the art, however many works were limited to univariate data, which consists of only a single input channel. Recent work has brought attention to Multivariate Time Series (MTS), where each example of data has multiple channels of information over time. However, MTS datasets are often small, with some training sets smaller than 100 examples, see \cite{bagnall2018uea}. This severely disadvantages deep learning methods, such as InceptionTime \cite{fawaz2020inceptiontime}, that have millions of parameters and can easily overfit on small datasets. 

Recent work by \cite{ruiz2021great} has shown that non-deep learning methods achieve higher MTS classification accuracy on a suite of benchmarks, citing these disadvantages. However, this work demonstrates that deep learning methods can outperform non deep methods with proper data augmentation. With the augmentations presented in this work, the InceptionTime network (introduced by \cite{fawaz2020inceptiontime}) exceeds the state of the art classification accuracy for 13 of the 26 benchmark datasets from \cite{bagnall2018uea} that were examined by \cite{ruiz2021great} and matches on another 2 datasets.

Comparison studies, such as the one by \cite{ruiz2021great}, often utilize multiple MTS datasets to compare multiple methods. When considering MTS classication methods for practical application, these studies may be biased towards methods that avoid overfitting on small datasets. By adding augmentations for deep networks, it is possible to reduce the bias and provide more accurate guidance for the application of methods onto real world problems. 

In this paper, we evaluate four MTS augmentations on the 26 benchmark datasets using three models. We show statistically significant improvements to classification accuracy when using augmentations. Additionally, we demonstrate that these four augmentations are applicable across a wide variety of dataset types and deep learning models. Furthermore, we provide theories as to why certain augmentations perform well on certain datasets and others do not. In addition to the above, we provide all the code necessary to reproduce the results and a Juptyer Notebook that anyone can run using Google Colab for replication\footnote{https://github.com/2022ICMLRAMTSC/CODE}. This repository was created anonymously For the double-blind review process. 

\section{Related Work}

\subsection{Time Series Augmentation}

Augmentation in time series data is a developing area of research. \cite{wen2020time} identifies 3 major time series tasks for data augmentation: classification, forecasting, and anomaly detection. They also note that augmentation methods can be broadly split into two categories, basic and advanced. The authors of this paper interpret basic augmentation methods as those that are applied to a batch without information outside of the batch being augmented, while advanced augmentation methods require outside of batch information. For example, a flip augmentation applied to a batch that is not influenced by any data outside of the batch should be considered basic. The Feature Space augmentation described by \cite{devries2017dataset} would be considered advanced, as it requires computing a feature space, which is computed outside of the batch. 

\begin{figure}[H]
    \centering
    \includegraphics[width=0.47\textwidth]{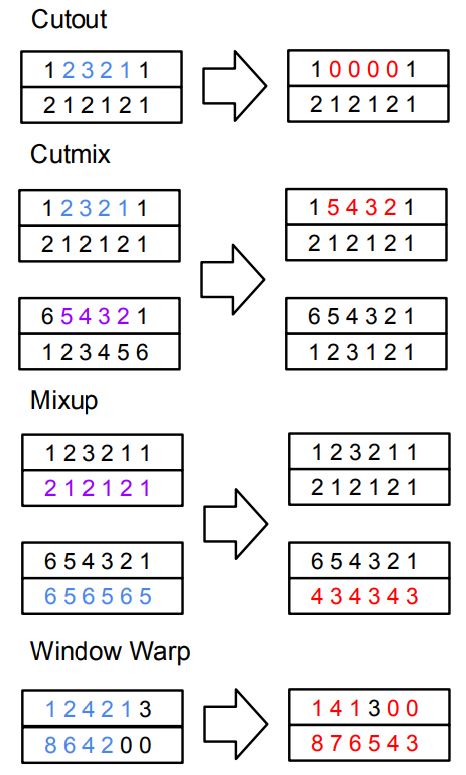}
    \caption{Examples of each augmentation examined in this paper in action. Blue indicates the targeted section of the time series and red indicates how the time series changed. For cutmix and mixup, purple indicates a reference from another time series, where the purple values are used in calculating the augmentation.}
    \label{fig:augs}
\end{figure}

While more detailed taxonomies of time series augmentations exist, see \cite{wen2020time} and \cite{iwana2021empirical}, the separation of basic and advanced is sufficient for this paper, which only deals with basic augmentations. The work by \cite{wen2020time} provides extremely limited empirical results and empirical results from \cite{iwana2021empirical} and \cite{le2016data} are limited to univariate datasets. There is a lack of extensive empirical evidence as to the effectiveness of augmentations on MTS datasets. 

\subsection{Image Augmentation} 

Numerous methods have been developed for the augmentation of image data, and with over 2000 citations recorded on Google Scholar, the survey of image augmentations by \cite{shorten2019survey} shows much greater interest than similar surveys in time series augmentation \cite{wen2020time} \cite{iwana2021empirical}. Before the popularization of transformers for vision by \cite{dosovitskiy2020image}, most computer vision tasks involved convolutional neural networks, with augmentations designed to improve the robustness of such networks. Similarly, many TSC networks utilize convolutions, suggesting that image augmentation methods may also apply to time series data.

\subsection{Time Series Classification}

Several methods have been developed for MTS classification, for a review see \cite{fawaz2019deep}. Neural network MTS classifiers tend to utilize some combination of Recurrent Neural Network (RNN) and Convolutional Neural Network (CNN) methods \cite{karim2017lstm}, or Temporal CNN (TCNN) methods  \cite{wang2017time}. Attention based models have also been used in time series, see \cite{song2018attend,yang2021predictive}. These methods involving deep neural networks require millions of parameters, which often leads to overfitting on smaller sized time series datasets. 

\section{Augmentations}

The augmentations examined in this work were implemented for TPU compilation, which limits the available operations in comparison to GPU and CPU augmentations. These augmentations had marginal impact on training time, which is expected because augmentation occurs before a batch of data is placed on the GPU or TPU. Figure \ref{fig:augs} shows an example of how the augmentations modify a time series. To the authors knowledge, window warping has not yet been investigated as an augmentation for time series data, and cutout, cutmix and mixup have not been mentioned in TSC literature prior to \cite{yang2021predictive}.

\subsection{Window Warp}

Window warping is a data augmentation technique that stretches or shrinks data in the temporal dimension, as described by \cite{le2016data}. After extracting a slice of the time series, the slice is resized using bilinear interpolation. The slice is reinserted into the same location, resulting in lengthened or shortened time series. This paper's implementation crops or pads (with zeros) the end of the sequence of the time series back to its original size, after window warping. Bilinear interpolation is implemented using Tensorflow's image resize function by reshaping the time series from a shape of $(l, c)$ to $(1, l, c)$. The position of the window is randomly determined. The size of the window is randomly determined and ranges between the max and min hyperparameter values. The scale factor is a hyper parameter, with 2.0 indicating a doubling of the length of the window. This operation applies across all channels.

\subsection{Cutout}

Cutout is an image augmentation technique that removes contiguous sections of data, as described by \cite{devries2017improved}. Temporal cutout selects a random time segment and set of channels from a MTS and sets selected values to 0. The size of the random time segment is randomly determined and ranges between the max and min hyperparameter values. The chance of each channel being selected for this operation is a hyperparameter called \emph{channel probability}.

\subsection{Mixup}

Mixup is an image augmentation technique that combines two examples of data, as described by \cite{zhang2017mixup}. Temporal mixup multiplies a randomly selected set of channels in the first MTS by a value, $m$, and then adds all values from the same set of channels from a randomly selected second MTS of any label multiplied by $1 - m$. The value of $m$ is randomly selected between a min and max hyperparameter. This operation applies to all channels. Note that this implementation does take the average of the two series' labels. 

\subsection{CutMix}

Cutmix is an image augmentation technique that combines two examples of data, as described by \cite{yun2019cutmix}. Temporal cutmix selects a random time segment from the first MTS and a random time segment of a random second MTS of any label. Note that the two random segments start and end at the same time step for both sequences. It then selects a random set of channels and replaces the first MTS's segment's channel values with those of the second. The size of the random time segment is randomly determined and ranges between the max and min hyperparameter values. The chance of each channel being selected for this operation is a hyperparameter called \emph{channel probability}.

\section{Datasets}

This paper evaluates the impact of the time series augmentations on the datasets in the UEA MTSC archive \cite{bagnall2018uea}. On release in 2018 it contained 30 multivariate datasets, of which four are not all equal length. We restrict our attention to the 26 equal length series to avoid any complications resulting from unequal length. An overview of the different datasets is provided in Appendix \ref{app:dataset}. Due to the page limits of this paper, we direct readers to \cite{bagnall2018uea} and \cite{ruiz2021great} for extended descriptions of the datasets.

\section{Models}

This paper explores three deep learning architectures that are often used for time series classification. By evaluating three different architectures, this paper hopes to demonstrate that these time series augmentations can be generalized for most deep learning architectures. 

\subsection{Convolutional Networks and InceptionTime}

Convolutional networks are very popular in the domain of computer vision, but can also perform well in time series classification. \cite{fawaz2020inceptiontime} proposed the InceptionTime model as an ensemble of five Inception models. Each Inception model is composed of two residual blocks, each containing three Inception modules, followed at the very end by a global average pooling layer and a dense classification head layer. Inception modules contain convolutions of various kernel sizes and a bottleneck layer. The residual blocks help mitigate the vanishing gradient issue by allowing for direct gradient flow \cite{he2016identity}. \cite{fawaz2020inceptiontime} noted that ensembling was necessary due to the high standard deviation in accuracy of single Inception models and the small size of time series datasets.

For this study, we evaluate the Inception model without ensembling, so references to InceptionTime refer to just the Inception model, without ensembling. There are two justifications for evaluating the model without ensembling. First, the compute cost is reduced five-fold allowing for a more efficient use of compute resources. Second, because ensembling can be considered as a regularization method on its own, it is better to analyze the regularization benefits provided by augmentations alone. 

% \begin{figure}[h]
%     \centering
%     \includegraphics[width=0.47\textwidth]{inceptiontime.JPG}
%     \caption{Inception model, used in the InceptionTime ensemble.}
%     \label{fig:itime}
% \end{figure}

\subsection{Recurrent Neural Networks}

\cite{husken2003recurrent} used a recurrent neural network (RNN) to process time series data for classification. Researchers have developed improved versions of RNN based networks for a variety of tasks, such as the Convolutional LSTM for classification, \cite{karim2017lstm}. However, RNN networks do not scale well with increasing sequence length, due to the vanishing gradient problem mentioned by \cite{le2016quantifying}. Additionally, due to their recurrent nature, RNNs must process a time series sequentially, whereas a convolutional network can apply the convolutions across the time series in parallel, which does not lend itself to effective use of GPUs and TPUs. This results in longer compute times during training. 

This paper implements a simple RNN network, refered to as simpleRNN. It consists of 3 LSTM layers of 256, 512, and 512 units. The final LSTM layer returns only the units at last time step, which is then used for classification in a dense layer with softmax. Due to the compute costs in long sequences, this paper evaluates 18 datasets in the UEA MTSC archive of 512 length or shorter for simpleRNN. 

\subsection{Self Attention Networks}

Multi-Headed Self Attention (MHSA) modules were popularized by \cite{devlin2018bert} for usage in Natural Language Processing and by \cite{dosovitskiy2020image} for Computer Vision. Research into MHSA for MTS classification is growing, as \cite{song2018attend} and \cite{russwurm2020self} use MHSA based networks for classification. While attention models do not suffer from vanishing gradients caused by long sequence lengths, the memory requirement increases by the square of the length of the sequence. This is because the scaled dot product attention that makes up the core of the MHSA architecture requires a dot product of two matrices along length of the time series, resulting in matrix shaped as \((batch size, length, length)\). 

This paper implements a simple MHSA network, refered to as simpleMHSA. It consists of a 1D convolutional layer with a kernel size of 1 and 512 filters, followed by a positional encoding layer. It then applies two encoder layers, with each encoder layer consisting of a single MHSA layer with 8 heads and 512 units and a feed forward layer. The output of the last time step of the second encoder layer is sent to a dense softmax classification layer. Due to the compute costs in long sequences, we evaluate 18 datasets in the UEA MTSC archive of 512 length or shorter for simpleMHSA. 

\subsection{Statistical Methods}

\cite{ruiz2021great} evaluates a wide variety of non deep learning methods on the UEA MTSC archive. For this paper, we will compare our results from the augmented deep learning networks with their results for both deep learning and non deep learning methods. Notable statistical methods that performed well on the UEA MTSC archive include ROCKET \cite{dempster2020rocket}, HIVE COTE \cite{lines2018time}, and CIF \cite{middlehurst2020canonical}.

\section{Experiments} 

\subsection{Augmentation Setup}

We evaluate different augmentation combinations, each with their own augmentation code, in order to measure the impact of individual augmentations both separately and when combined with other augmentations. It should be noted that there are an infinite number of possible combinations to evaluate  (we can evaluate every value of P between 0 and 1, the probability of applying the augmentation). We limit our search to eight augmentation pipelines, four of which evaluate the four augmentations in isolation, three of which evaluate augmentations together, and one control augmentation, which does not modify the data in any way. It is important to note that there are an infinite number of possible combinations for augmentations. The specific combinations of augmentations are arbitrary, except for the isolated augmentations, which were designed to evaluate the augmentations in isolation. 

Across all augmentation combinations, certain input parameters are dynamically defined based on the dataset. For Mixup, there are no input shape dependent parameters. For cutout and cutmix, given $l_{max}$ as the maximum length of the input time sequences, the segment to be cutmixed or cutout has a length $s_{length}$ selected uniformly at random, $s_{length} \sim U(\frac{1}{2}l_{max},l_{max})$, starting at a position, $p$, selected uniformly at random, $p \sim U(0, l_{max} - s_{length} - 1)$, where 0 is the starting index of the time series. For window warping, the segment to be lengthened or shorted has a length $s_{length}$ selected uniformly at random, $s_{length} \sim U(\frac{1}{8}l_{max}, \frac{1}{3}l_{max})$, with a starting position, $p$ selected similarly, uniformly at random, $p \sim U(0, l_{max} - s_{length} - 1)$. The new length of the segment is the old length multiplied by a static scaling hyper parameter $S$. See Table \ref{table:augs} for an explanation of static hyper parameters for each augmentation. 

%For cutmix, the segment to be cut and mix is of a random size between the maximum cutmix length and minimum cutmix length. The maximum cutmix length is set to the dataset time sequence max length and the minimum cutmix length is set to half of that. For cutout, the segment to be cut out is of a random size between the maximum cutout length and minimum cutout length. The maximum cutout length is set to the dataset time sequence max length and the minimum cutout length is set to half of that. For window warp, the segment to be lengthened or shortened is of a random size between the maximum window size and minimum window size. The maximum window size is set to one third of the sequence max length and the minimum window size is set to one eight of sequence max length. 

\begin{table}[t!]
  \centering
\begin{tabular}{|l|l|}
\toprule
Aug Code & Definition  \\
\midrule
A     &   Cutmix(P = 0.8, CP = 0.5)\\ 
                        & Cutout(P = 0.8, CP = 0.5) \\
                        & Mixup(P = 0.8)   \\ \hline
B & Cutmix(P = 0.8, CP = 0.2)   \\
    & Cutmix(P = 0.8, CP = 0.2)   \\
    & Cutout(P = 0.8, CP = 0.2)   \\
    & Mixup(P = 0.8)   \\ \hline
C  & Cutout(P = 0.5, CP = 0.3)   \\
   & Cutout(P = 0.5, CP = 0.3)  \\ \hline
D  & Mixup(P = 0.5)  \\
   & Mixup(P = 0.5)  \\ \hline
E  & Cutmix(P = 0.5, CP = 0.3)  \\
    & Cutmix(P = 0.5, CP = 0.3)  \\ \hline
F  &  Windowwarp(P = 0.5, S = 0.5) \\
    &  Windowwarp(P = 0.5, S = 2.0) \\ \hline
G  & Cutmix(P = 0.8, CP = 0.2)   \\
    & Cutout(P = 0.8, CP = 0.2)   \\
    & Cutout(P = 0.8, CP = 0.2)   \\
    & Mixup(P = 0.8)   \\  
    &  Windowwarp(P = 0.5, S = 0.5) \\
    &  Windowwarp(P = 0.5, S = 2.0) \\ 
\bottomrule
\end{tabular}
\caption{Augmentation Combinations, with their associated code and static hyper parameter values. For each augmentation code, augmentations are applied from top to bottom in order. P represents the probability to apply the augmentation. CP represents the probability for each channel to be selected for the augmentation and if not present indicates the augmentation applies to all channels. S presents the scale factor for window warping.}
\label{table:augs}
\end{table}

\subsection{Experimental Setup}

All experimental results were generated using Google Colab instances with a v2-8 TPU. In each training run, models were trained for up to 1500 epochs using a batch size of 64. If a model did not improve its validation loss for 100 epochs, the training process was stopped. Models used an Adam optimizer, with an initial learning rate of 1e-4, which was cut in half if the model's loss did not improve for 10 epochs. Each epoch was 100 batches of data, or enough batches for one full pass through the training set, whichever was greater. This was necessary to prevent epochs where the model trains for only one batch for certain datasets. All models were trained with categorical crossentropy loss. 

We define one experiment as a single training run for a single combination of neural network model, dataset, and augmentation pipeline. We repeat the experiment for each combination using 29 different splits of the dataset and once with the original train test split of the dataset. The 29 splits were generated randomly, but with a preset seed value. This ensures that each single split is the same across models and augmentation pipelines. When generating splits, we combined the train and test datasets and resampled new train and test datasets, such that the number of examples for each class in the new split matches the original train and test split. 

To compute statistical significance, we use the two sample T test, with the assumption of unequal variances, to compare the best accuracy. We consider an augmentation to be statistically better/worse than no augmentations if it passes a two sided T test with $a = 0.05$ and its mean accuracy is greater/lesser than without augmentations. 

Due to memory constraints, we only evaluated the simpleRNN and simpleMHSA networks on datasets with 512 or less time steps. Due to limited resources, SimpleRNN and simpleMHSA experiments are repeated for a total of 10 times, one original train test split and 9 generated splits. 

\begin{table*}[t]
  \centering
\begin{tabular}{llrrrrrrrr}
\toprule
    & Aug Code &       None &      A &      B &      C &      D &      E &      F &      G \\
Model & Dataset &         &        &        &        &        &        &        &        \\
\midrule
IT & ArticularyWordRecognition &   99.52 & \textit{-1.82} &  -0.18 &  -0.07 & \textbf{0.19} & \textit{-0.89} &   0.02 & \textbf{0.22} \\
    & AtrialFibrillation &   39.33 & \textbf{7.11} & \textbf{7.33} & \textbf{5.56} &   0.67 & \textbf{5.33} &   0.00 & \textbf{6.44} \\
    & BasicMotions &  100.00 & \textit{-0.33} &  -0.17 &   0.00 &   0.00 &  -0.08 &   0.00 &   0.00 \\
    & Cricket &   99.95 &  -0.09 &  -0.09 & \textit{-0.32} &   0.05 &  -0.23 &   0.00 &  -0.09 \\
    & DuckDuckGeese &   69.53 &  -1.33 &  -0.60 & \textbf{7.20} & \textit{-3.87} &  -0.13 &   1.40 & \textbf{5.13} \\
    & ERing &   95.02 & \textit{-3.96} & \textit{-1.40} & \textbf{2.14} & \textbf{1.07} & \textit{-3.52} & \textit{-1.37} & \textbf{0.96} \\
    & EigenWorms &   76.72 & \textbf{14.83} & \textbf{13.97} &   0.66 & \textbf{10.53} & \textit{-8.24} &   2.60 & \textbf{14.73} \\
    & Epilepsy &   99.69 & \textit{-1.47} & \textit{-0.63} &   0.14 &  -0.12 & \textit{-1.52} & \textbf{0.24} & \textbf{0.24} \\
    & EthanolConcentration &   33.52 & \textbf{47.96} & \textbf{43.27} & \textbf{14.75} & \textbf{4.71} & \textbf{38.75} &  -0.32 & \textbf{10.56} \\
    & FaceDetection &   71.08 & \textbf{1.20} & \textbf{2.19} &   0.23 &  -0.50 &   0.66 & \textit{-5.07} &  -0.30 \\
    & FingerMovements &   63.03 &  -1.50 &  -1.70 & \textit{-3.13} & \textbf{2.73} & \textit{-2.80} &   0.50 & \textit{-2.73} \\
    & HandMovementDirection &   49.73 &  -1.44 &   0.41 &  -0.59 &   1.62 &  -0.50 &  -1.53 & \textit{-2.21} \\
    & Handwriting &   63.36 & \textit{-15.60} & \textit{-8.48} & \textit{-6.68} & \textit{-3.05} & \textit{-10.10} & \textbf{3.82} & \textit{-3.76} \\
    & Heartbeat &   77.64 &   0.13 & \textbf{0.94} &   0.62 &  -0.60 &   0.08 &  -0.02 & \textbf{1.40} \\
    & LSST &   49.88 &   0.11 & \textit{-6.10} &  -1.53 &   2.01 & \textit{-8.82} &  -1.11 &  -0.82 \\
    & Libras &   92.13 & \textit{-1.70} & \textit{-0.85} &  -0.41 &   0.50 & \textit{-4.20} & \textbf{4.04} & \textbf{4.76} \\
    & MotorImagery &   58.80 &   1.10 &   1.50 &   1.13 &   1.27 &   1.03 &   0.90 &  -0.17 \\
    & NATOPS &   97.83 & \textit{-8.50} & \textit{-7.54} & \textit{-2.41} &   0.20 & \textit{-6.37} & \textbf{0.96} & \textit{-1.11} \\
    & PEMS-SF &   85.51 &  -0.73 & \textbf{2.76} & \textbf{3.28} &   0.92 & \textbf{3.85} &  -1.48 & \textbf{1.97} \\
    & PenDigits &   99.75 & \textit{-0.96} & \textit{-0.51} &  -0.04 &  -0.02 & \textit{-0.60} & \textit{-0.11} & \textit{-1.04} \\
    & PhonemeSpectra &   30.71 & \textit{-2.41} &  -0.17 & \textbf{1.56} &  -0.10 & \textit{-0.50} & \textbf{1.35} & \textbf{2.84} \\
    & RacketSports &   93.77 & \textit{-4.52} & \textit{-1.54} &  -0.09 &   0.07 & \textit{-1.75} & \textbf{1.56} & \textbf{1.78} \\
    & SelfRegulationSCP1 &   85.93 & \textit{-2.54} &   0.55 & \textit{-1.67} & \textbf{1.89} & \textit{-1.31} &  -0.56 & \textbf{1.99} \\
    & SelfRegulationSCP2 &   57.04 & \textbf{2.06} & \textbf{1.22} & \textbf{1.89} &   0.76 & \textbf{1.78} &  -0.30 & \textbf{1.44} \\
    & StandWalkJump &   52.22 &   4.22 &   4.22 & \textbf{6.67} &  -0.67 & \textbf{6.44} &  -1.11 & \textbf{4.89} \\
    & UWaveGestureLibrary &   93.61 & \textit{-3.66} & \textit{-1.10} &  -0.55 & \textbf{1.59} & \textit{-4.40} &  -0.23 & \textbf{1.33} \\
\bottomrule
\end{tabular}

\caption{Difference in accuracy for the InceptionTime model between no Augmentation and the Specific Augmentation Combination for the Inception Time Model. Entries are bold where the difference is statistically significant and the Augmentation Combination's accuracy is greater. Entries are italicized where the difference is statistically significant and the Augmentation Combination's accuracy is lower.}
\label{table:ITIME}
\end{table*}

\section{Discussion}

See Table \ref{table:ITIME} for experimental results for InceptionTime and Appendix \ref{app:rnnmhsa} for RNN and Self Attention. 

\subsection{Overfitting}

When training deep neural networks with small datasets, the biggest challenge is often generalization from the training set to the testing set. When there is no augmentation, we observe this problem with the InceptionTime model in almost all datasets. For example, when training on the EthanolConcentration dataset without augmentation, the best test accuracy is achieved in the first epoch, at approximately 40\% train accuracy and 30\% test accuracy. In future epochs, the train accuracy increases and the test accuracy decreases, resulting in approximately 95\% train accuracy and 25\% test accuracy by the 50th epoch and an early stopping of the training process at the 100th epoch. 

This is not the case when we use augmentations. With Augmentation Combination B, we will often observe a train accuracy of around 65\% and a test accuracy of 70\% at the 300th epoch. It's quite possible that with very aggressive augmentations, the model may suffer from underfitting, where the model has difficulty learning the training data. Regardless, the use of augmentations is one method for reducing overfitting and improving the model's ability to generalize. 

\subsection{Analysis of Augmentations and Datasets}

\subsubsection{Cutout}

Augmentation combination C isolates the cutout augmentation. Compared other isolated augmentation combinations, cutout appears to perform well in the spectrogram datasets, which are DuckDuckGees, Heartbeat, PhonemeSpectra, and StandWalkJump. Because a spectrogram is a visual representation of the spectrum of frequencies of a signal as it varies with time, it is realistic for for certain frequencies to be absent at certain time steps. The cutout augmentation does exactly that, setting random channels (frequencies) for a random segment of the time series to zero. 

We would expect cutout to perform poorly in datasets where zero values are unrealistic, which is what we observe for the Handwriting dataset. The Handwriting dataset records a subject's writing hand's accelerometer data as they write letters of the alphabet. It is possible that zero values in this case only confuse the model during training. However, we observe significant improvement for SelfRegulationSCP2, which is an Electroencephalogram (EEG) dataset. Because EEG data normally oscillates as a wave, it is unusual that randomly setting long segments of EEG data to zero would benefit the model. 

Based on the above analysis, further research can be done to improve or modify cutout for non spectrogram data. For example, rather than setting the values of a time segment to zero, one could set the values to the mean of that channel. This could result in a version of cutout that sets all frequencies to zero. Alternatively, one could transform a time series into a spectrogram, perform frequency cutout, and then transform the result back. This could be used to cut specific frequencies. 

\subsubsection{Mixup}

The EigenWorms dataset measures the movement of various roundworms on an agar plate according to 6 eigen vectors, with the goal of classifying the individual worm as either a wild type or one of 4 mutant types. The mixup augmentation combination D performs best out of the 4 isolated augmentations. It is plausible that mixup generates time sequences that are more realistic or plausible when compared to cutmix. One would expect mixup to generate intermediary movement patterns, where the time sequence adopts some characteristics of another one. This should result in a more diverse training set of data.

\subsubsection{Cutmix}

Augmentation combination E isolates the cutmix augmentation, which performs extremely well on the EthanolConcentration dataset. \cite{ruiz2021great} performed a case study on why deep learning methods struggle so much with this dataset. In summary, the discriminatory features lie in a small region and that the variation in this small region is significantly smaller than the rest of the sequence. When these factors are combined with a relatively small training set size (261 examples), they observe that non augmented deep learning models tend to focus on irrelevant regions of the sequence. Cutmix directly addresses this issue by swapping regions between examples, regardless of class. Given 4 classes distributed equally and if we assume that 10\% of the sequence is relevant for prediction, there is a 90\% chance of swapping noise between two examples and a 67\% chance of swapping noise between examples of different classes. This will strongly penalize any model from learning spurious features from the noise in the sequence. While the percentage chance estimates are not completely accurate, it should be clear that one should expect cutmix performs quite well when the discriminatory features lie in a small region. 

Another case where cutmix can perform well is when cutmix results in plausible transformations of the time seqeuence. In the PEMS-SF dataset, each example tracks traffic data in 10 minute intervals for 1 day, with the goal of classifying the day of week. Swaps between two examples can generate plausible traffic patterns and help reduce overfitting on the training data. 

Overall, we tend to observe that cutmix performs well on the same datasets where cutout performs well. This is likely because both augmentations remove a section of the original time series, with cutmix replacing that section. However, cutmix is the augmentation most likely to reduce accuracy by more than 3\%, which it does for 6 datasets. 

\subsubsection{Window Warp}

Augmentation combination F isolates the window warp augmentation, which performs much better than other augmentations on the Handwriting and Libras dataset. The Handwriting dataset measures accelerometer data from the writing hand as an individual writes letters of the alphabet, while Libras measures the 2D coordinate of gestures from Brazilian sign language. We also observe noticeable improvements in the RacketSports dataset, which measures gyroscope and accelerometer data. These 3 datasets measure movements executed by people, in which the execution speed can vary significantly between individuals. In comparison with other augmentations, window warp rarely harms classification performance by a significant amount.

\subsubsection{Combining Augmentations Together}

When we compare augmentations combinations solely on the basis whether or not it improves performance, combination G improves performance in 16 datasets out of 26. However, it does not always improve performance by the largest amount. For the Handwriting dataset, using all 4 augmentations together performs much worse than using just the window warping augmentation. Because the nature of time series data can vary significantly, different augmentations can have drastically different effects. It is important to evaluate a wide variety of augmentations both individually and together. 

Furthermore, it is important to evaluate different hyper parameter values for the augmentations. Augmentation combinations A and B are almost identical, but B applies cutmix twice and applies all of its augmentations with a lower channel probability. Some datasets, such as the PEMS-SF dataset benefit more from combination B than combination A. 

Unfortunately, finding the best combination of augmentations may require a very thorough cross validation process. The authors would recommend beginning such a process with isolated augmentations, with the parameters specified in this paper. Doing so may reveal which type of augmentations perform best and provide more insight on what may work for that particular dataset.  

\subsubsection{Effects on RNN and Self Attention Networks}

Results on the simpleRNN and simpleMHSA networks are limited to shorter datasets and lower statistical power due to a smaller sample size, with results presented in Appendix \ref{app:rnnmhsa}. However, the results show quite a different trend compared to the InceptionTime results. In particular, the window warp and mixup augmentations provide a statistically significant improvement far more often in these two networks than in InceptionTime. It is likely that architectural differences create different biases in networks, resulting in networks learning features in different ways and interacting with augmentations differently. 

One potential source of difference is the use of the Global Average Pooling layer in the InceptionTime model, which is not present in both the simpleRNN and simpleMHSA model. Without Global Average Pooling, only the output of the last time step is used for predicting the class. Global Average Pooling can also ignore empty sections or swapped sections of the time series. This could explain why the simpleMHSA does not benefit much from cutout and cutmix. 

Regardless, the results show that most augmentations provide a benefit for the simpleRNN model, which suggests that more complex RNN based models may also benefit from these augmentations. The simpleMHSA model appears to benefit only from mixup and window warping. Further research into the impact of augmentations on these model architectures is needed. 

\subsection{Comparison with Other Methods}

For each dataset, we compare the best augmentation combination for the InceptionTime model with the best model found in the work by \cite{ruiz2021great}, see Table \ref{table:compare}.  If we exclude datasets where our accuracy and the state-of-the-art best results from \cite{ruiz2021great} are both 100\%, we show that at least one augmentation combination for InceptionTime performs better in 13 of 24 datasets compared with their best model. The augmentations allow for the InceptionTime network to outperform non deep learning methods, as well as the 5 network ensembled version of InceptionTime. 

The work by \cite{ruiz2021great} suggests that non deep learning methods, such as ROCKET, CIF, and HIVE-COTE, are more effective at MTS classification than deep learning methods. However, deep learning methods are prone to overfitting and the UEA MTSC Archive contains a large number of datasets with fewer than 300 training examples. We believe that future comparisons between methods should include datasets with more training examples and evaluations of models with and without augmentations.

\begin{table}[t!]
  \centering

\begin{tabular}{lrrl}
\toprule
{} &  Our Acc & Ruiz Acc & Algorithm \\
Dataset Name             &                                  &          &           \\
\midrule
 \textbf{AWR} & \textbf{99.74} & \textbf{99.56} & \textbf{ROCKET} \\
AtrialFibrillation        &                            46.67 &    74.00 &      MUSE \\
 \textbf{BasicMotions} & \textbf{100.00} & \textbf{100.00} & \textbf{Multiple} \\
 \textbf{Cricket} & \textbf{100.00} & \textbf{100.00} & \textbf{Multiple} \\
 \textbf{DuckDuckGeese} & \textbf{76.73} & \textbf{63.47} & \textbf{IT} \\
ERing                     &                            97.16 &    98.05 &    ROCKET \\
 \textbf{EigenWorms} & \textbf{91.55} & \textbf{90.33} & \textbf{CIF} \\
Epilepsy                  &                            99.93 &   100.00 &        HC \\
EC      &                            81.48 &    82.36 &       STC \\
FaceDetection             &                            73.27 &    77.24 &        IT \\
 \textbf{FingerMovements} & \textbf{65.77} & \textbf{56.13} & \textbf{IT} \\
HMD    &                            51.35 &    52.21 &       CIF \\
 \textbf{Handwriting} & \textbf{67.19} & \textbf{65.74} & \textbf{IT} \\
 \textbf{Heartbeat} & \textbf{79.04} & \textbf{76.52} & \textbf{CIF} \\
LSST                      &                            51.88 &    63.62 &      MUSE \\
 \textbf{Libras} & \textbf{96.89} & \textbf{94.11} & \textbf{ResNet} \\
 \textbf{MotorImagery} & \textbf{60.30} & \textbf{53.80} & \textbf{TSF} \\
 \textbf{NATOPS} & \textbf{98.80} & \textbf{97.11} & \textbf{ResNet} \\
PEMS-SF                   &                            89.36 &    99.68 &       CIF \\
PenDigits                 &                            99.73 &    99.85 &        IT \\
PhonemeSpectra            &                            33.55 &    36.74 &        IT \\
 \textbf{RacketSports} & \textbf{95.55} & \textbf{92.79} & \textbf{ROCKET} \\
SCP1        &                            87.92 &    95.68 &    TapNet \\
 \textbf{SCP2} & \textbf{59.09} & \textbf{53.69} & \textbf{DTW} \\
 \textbf{StandWalkJump} & \textbf{58.89} & \textbf{45.56} & \textbf{ROCKET} \\
 \textbf{UW} & \textbf{95.21} & \textbf{94.43} & \textbf{ROCKET} \\
\bottomrule
\end{tabular}

\caption{The mean of the accuracies across folds of the best augmentation combination for each dataset compared with the best results reported by \cite{ruiz2021great} (see Table 10 of their work). Bold rows indicate better or equal accuracy. Dataset names were shortened for formatting. ArticularyWordRecognition(AWR), EthanolConcentration(EC), HandMovementDirection(HMD),  SelfRegulationSCP1(SCP1),  SelfRegulationSCP2(SCP2), UWaveGestureLibrary(UW)  }

\label{table:compare}
\end{table}

\subsection{Future Work}

Augmentations are normally applied to deep learning networks, but there may be a benefit to applying augmentations to non deep methods, such as ROCKET, CIF, and HIVE-COTE. This is an area research that could be further explored, as the work by \cite{ruiz2021great} shows how these methods can outperform deep learning methods.

Time series forecasting can also benefit from augmentation, as shown by \cite{bandara2021improving}. They evaluate 3 advanced augmentation methods and do not evaluate any of the methods in this paper. Further work can be done to evaluate the basic methods in this paper for the time series forecasting task. 

Further development of deep learning methods for time series may be limited by the overfitting issue, due to the small size of available datasets. With augmentations, we can develop complex models that would otherwise overfit.

Follow up work should evaluate the augmentations in this paper on more state of the art deep neural networks. Due to the resource limits, the authors of this paper could not evaluate the ensembled InceptionTime, but such an evaluation could prove quite helpful in understanding how ensembling and augmentations interact. 

\section{Conclusion}

The experimental results above demonstrate that the InceptionTime network, with augmentations and without ensembling, is the best classifier for the UEA MTSC Archive; it matches or exceeds classification accuracy on 15 of the 26 datasets. The improvement brought about by augmentations can be as drastic as a 48\% increase in accuracy from 33\% to 81\% on the Ethanol Concentration dataset. These improvements suggest that deep neural networks are strongly overfitting on MTS datasets. 

We recommend that future researchers and engineers fully leverage the benefits of augmentation to improve performance on classification tasks. While it does require more time, finding the right augmentation combination can easily and significantly improve performance. With augmentations, we can design and train deeper models with a lowered risk of overfitting.

However, the use of augmentations can complicate the comparison between models. Could model A perform better than model B using different hyper parameters for the augmentations? By recommending augmentations for comparison studies, are we opening a Pandora's box of infinite combinations for evaluation? The authors of this paper notes that the box was already open; there's already an infinite combination of learning rates, optimizers, schedulers, and many more hyper parameters. The key contribution of this work is to add impactful augmentations for designing, evaluating, and applying deep MTS classification models when overfitting is a problem, which it always is. 

% % Acknowledgements should only appear in the accepted version.
% \section*{Acknowledgements}

% \textbf{Do not} include acknowledgements in the initial version of
% the paper submitted for blind review.

% If a paper is accepted, the final camera-ready version can (and
% probably should) include acknowledgements. In this case, please
% place such acknowledgements in an unnumbered section at the
% end of the paper. Typically, this will include thanks to reviewers
% who gave useful comments, to colleagues who contributed to the ideas,
% and to funding agencies and corporate sponsors that provided financial

\nocite{langley00}

\bibliography{example_paper}
\bibliographystyle{icml2022}

%%%%%%%%%%%%%%%%%%%%%%%%%%%%%%%%%%%%%%%%%%%%%%%%%%%%%%%%%%%%%%%%%%%%%%%%%%%%%%%
%%%%%%%%%%%%%%%%%%%%%%%%%%%%%%%%%%%%%%%%%%%%%%%%%%%%%%%%%%%%%%%%%%%%%%%%%%%%%%%
% APPENDIX
%%%%%%%%%%%%%%%%%%%%%%%%%%%%%%%%%%%%%%%%%%%%%%%%%%%%%%%%%%%%%%%%%%%%%%%%%%%%%%%
%%%%%%%%%%%%%%%%%%%%%%%%%%%%%%%%%%%%%%%%%%%%%%%%%%%%%%%%%%%%%%%%%%%%%%%%%%%%%%%
\newpage
\appendix
\onecolumn
\section{Overview of UEA MTSC Archive Datasets }
\label{app:dataset}

\begin{table*}[h]
  \centering
\begin{tabular}{llrrrrr}
\hline Code & Name & Train size & Test size & Dims & Length & Classes \\
\hline AWR & ArticularyWordRecognition & 275 & 300 & 9 & 144 & 25 \\
AF & AtrialFibrillation & 15 & 15 & 2 & 640 & 3 \\
BM & BasicMotions & 40 & 40 & 6 & 100 & 4 \\
CR & Cricket & 108 & 72 & 6 & 1197 & 12 \\
DDG & DuckDuckGeese & 50 & 50 & 1345 & 270 & 5 \\
EW & EigenWorms & 128 & 131 & 6 & 17,984 & 5 \\
EP & Epilepsy & 137 & 138 & 3 & 206 & 4 \\
EC & EthanolConcentration & 261 & 263 & 3 & 1751 & 4 \\
ER & ERing & 30 & 270 & 4 & 65 & 6 \\
FD & FaceDetection & 5890 & 3524 & 144 & 62 & 2 \\
FM & FingerMovements  & 316 & 100 & 28 & 50 & 2 \\
HMD & HandMovementDirection & 160 & 74 & 10 & 400 & 4 \\
HW & Handwriting & 150 & 850 & 3 & 152 & 26 \\
HB & Heartbeat & 204 & 205 & 61 & 405 & 2 \\
LIB & Libras & 180 & 180 & 2 & 45 & 15 \\
LSST & LSST & 2459 & 2466 & 6 & 36 & 14 \\
MI & MotorImagery & 278 & 100 & 64 & 3000 & 2 \\
NATO & NATOPS & 180 & 180 & 24 & 51 & 6 \\
PD & PenDigits & 7494 & 3498 & 2 & 8 & 10 \\
PEMS & PEMS-SF & 267 & 173 & 963 & 144 & 7 \\
PS & PhonemeSpectra & 3315 & 3353 & 11 & 217 & 39 \\
RS & RacketSports & 151 & 152 & 6 & 30 & 4 \\
SRS1 & SelfRegulationSCP1 & 268 & 293 & 6 & 896 & 2 \\
SRS2 & SelfRegulationSCP2 & 200 & 180 & 7 & 1152 & 2 \\
SWJ & StandWalkJump  & 12 & 15 & 4 & 2500 & 3 \\
UW & UWaveGestureLibrary & 120 & 320 & 3 & 315 & 8 \\

\hline
\end{tabular}
\caption{UEA MTSC Archive Dataset Summary Stats}
\label{table:datasets}
\end{table*}

\newpage
\section{RNN and Self Attention Results}
\label{app:rnnmhsa}

\begin{table*}[h]
  \centering

\begin{tabular}{llrrrrrrrr}
\toprule
    & aug\_ver &       0 &      1 &      2 &      3 &      4 &      5 &      6 &      8 \\
model & dataset\_name &         &        &        &        &        &        &        &        \\
\midrule
MHSA & ArticularyWordRecognition &   98.28 &   0.56 & \textbf{0.56} & \textbf{0.52} & \textbf{1.26} &  -0.41 &   0.20 &   0.43 \\
    & BasicMotions &   99.75 & -10.37 & \textit{-3.57} &  -0.75 &   0.25 & \textit{-3.75} &   0.25 &   0.25 \\
    & DuckDuckGeese &   69.87 &  -5.87 & \textit{-7.66} & \textbf{4.73} & \textit{-8.67} &   2.53 &   1.02 &   2.38 \\
    & ERing &   92.72 &  -1.33 &   0.05 & \textbf{1.32} & \textbf{2.95} &  -1.23 & \textit{-2.14} &  -0.59 \\
    & Epilepsy &   92.15 &  -0.12 & \textbf{1.75} & \textbf{2.85} & \textbf{3.43} &  -1.57 & \textbf{4.79} & \textbf{3.86} \\
    & FaceDetection &   71.49 & \textit{-7.06} & \textit{-3.22} & \textit{-1.09} &  -0.49 & \textit{-3.06} & \textit{-6.03} & \textit{-16.04} \\
    & FingerMovements &   58.07 & \textbf{3.43} & \textbf{3.62} & \textbf{5.33} & \textbf{5.23} &   2.93 & \textbf{3.27} &   2.93 \\
    & HandMovementDirection &   36.80 & \textbf{10.83} & \textbf{18.46} & \textbf{9.01} & \textbf{28.87} & \textbf{7.52} & \textbf{5.09} & \textbf{6.78} \\
    & Handwriting &   27.59 & \textit{-3.88} &  -0.72 &   0.60 & \textbf{4.18} &   0.22 & \textbf{5.25} &   1.24 \\
    & Heartbeat &   73.09 &   1.42 & \textbf{1.34} &  -0.31 &   1.15 &   0.67 &   0.95 &   0.81 \\
    & LSST &   61.99 & \textit{-9.75} & \textit{-5.31} & \textit{-1.33} & \textit{-1.11} & \textit{-3.47} & \textbf{2.46} & \textbf{1.10} \\
    & Libras &   85.13 & \textbf{2.65} & \textbf{2.88} & \textbf{2.76} & \textbf{5.15} &  -1.13 & \textbf{7.28} & \textbf{6.40} \\
    & NATOPS &   93.26 & \textit{-7.15} & \textit{-6.86} & \textit{-4.31} & \textbf{3.96} & \textit{-5.93} & \textbf{5.07} & \textbf{2.99} \\
    & PEMS-SF &   89.83 & \textbf{2.08} &   0.41 &   0.17 &   0.46 &   2.31 &  -1.32 & \textit{-3.34} \\
    & PenDigits &   99.42 & \textit{-1.97} & \textit{-1.01} &  -0.24 &   0.07 &  -1.12 & \textit{-0.50} & \textit{-3.44} \\
    & PhonemeSpectra &   11.98 & \textbf{0.76} & \textbf{1.08} &   0.06 & \textbf{1.12} & \textbf{0.62} & \textbf{3.00} & \textbf{4.48} \\
    & RacketSports &   90.70 & \textit{-6.82} &  -1.16 &   0.29 & \textbf{2.52} & \textit{-2.68} & \textbf{2.35} & \textbf{2.88} \\
    & UWaveGestureLibrary &   85.90 &  -1.68 & \textbf{1.65} & \textbf{2.82} & \textbf{5.85} &  -2.52 & \textbf{4.90} & \textbf{5.43} \\
RNN & ArticularyWordRecognition &   94.78 &  -0.15 & \textbf{2.58} & \textbf{2.18} & \textbf{4.28} & \textbf{1.10} & \textbf{3.32} & \textbf{4.80} \\
    & BasicMotions &   99.08 &  -1.58 &   0.38 & \textbf{0.92} & \textbf{0.92} &  -1.00 & \textbf{0.92} & \textbf{0.92} \\
    & DuckDuckGeese &   60.93 & \textit{-9.18} & \textit{-9.15} &   2.19 & \textit{-7.77} &  -2.93 &   3.92 & \textbf{9.98} \\
    & ERing &   84.75 & \textbf{5.02} & \textbf{6.66} & \textbf{7.35} & \textbf{8.55} & \textbf{5.97} & \textbf{3.40} & \textbf{9.86} \\
    & Epilepsy &   88.43 & \textbf{6.23} & \textbf{6.99} & \textbf{7.40} & \textbf{8.85} & \textbf{4.58} & \textbf{4.84} & \textbf{11.37} \\
    & FaceDetection &   70.31 &  -1.43 & \textbf{0.85} &  -0.33 &  -0.12 &   0.53 & \textit{-4.64} & \textit{-3.99} \\
    & FingerMovements &   56.90 & \textbf{4.10} & \textbf{4.60} & \textbf{4.54} & \textbf{8.27} & \textbf{4.39} & \textbf{3.24} & \textbf{4.74} \\
    & HandMovementDirection &   39.55 & \textbf{7.07} & \textbf{7.55} & \textbf{5.55} & \textbf{13.49} &   2.42 & \textbf{4.85} & \textbf{13.28} \\
    & Handwriting &   20.93 & \textbf{4.08} & \textbf{20.05} & \textbf{4.81} & \textbf{40.61} & \textit{-1.72} & \textbf{23.42} & \textbf{30.20} \\
    & Heartbeat &   73.66 &   0.67 & \textbf{0.87} &   0.18 &  -0.33 & \textbf{0.80} & \textbf{3.62} & \textbf{5.90} \\
    & LSST &   65.15 & \textit{-1.85} & \textbf{1.96} & \textbf{2.67} & \textit{-0.49} & \textbf{2.43} & \textbf{1.96} & \textbf{4.67} \\
    & Libras &   74.41 & \textbf{15.94} & \textbf{16.94} & \textbf{9.24} & \textbf{15.41} & \textbf{9.61} & \textbf{12.26} & \textbf{22.06} \\
    & NATOPS &   94.76 & \textit{-3.51} & \textit{-3.47} &   0.87 & \textbf{4.13} & \textit{-1.82} & \textbf{3.73} & \textbf{3.22} \\
    & PEMS-SF &   84.97 & \textbf{2.96} & \textbf{3.30} & \textbf{4.01} & \textbf{2.31} & \textbf{6.46} &   1.16 & \textbf{2.15} \\
    & PenDigits &   99.53 & \textit{-1.73} & \textit{-0.39} &   0.02 & \textbf{0.16} & \textit{-0.51} &  -0.27 & \textit{-1.49} \\
    & PhonemeSpectra &   19.11 & \textbf{0.65} & \textbf{2.93} & \textit{-0.53} & \textbf{2.44} & \textit{-1.65} & \textbf{6.54} & \textbf{11.64} \\
    & RacketSports &   88.68 &   1.78 & \textbf{5.32} & \textbf{5.76} & \textbf{4.79} & \textbf{5.32} & \textbf{5.49} & \textbf{6.89} \\
    & UWaveGestureLibrary &   76.26 &   1.94 & \textbf{8.72} & \textbf{3.47} & \textbf{18.01} & \textit{-3.14} & \textbf{13.52} & \textbf{19.19} \\
\bottomrule
\end{tabular}

\caption{Results for RNN and Self Attention (MHSA)}
\label{table:rnnsa}
\end{table*}

% \section{Window Warp Pseudo Code}

% \section{Cutout Pseudo Code} 

% \section{Cutmix Pseudo Code} 

% \section{Mixup Pseudo Code} 

\end{document}